\documentclass{article}

\usepackage[final]{neurips_distshift_2021}

\usepackage[utf8]{inputenc} 
\usepackage[T1]{fontenc}    
\usepackage{hyperref}       
\usepackage{url}            
\usepackage{booktabs}       
\usepackage{amsfonts}       
\usepackage{nicefrac}       
\usepackage{microtype}      
\usepackage{xcolor}         

\usepackage{amsmath, amssymb}
\usepackage{natbib}
\usepackage{algorithm}
\usepackage{algpseudocode}
\usepackage{graphicx}
\usepackage{todonotes}

\DeclareMathOperator{\argmax}{arg\,max}
\newcommand{\R}{\mathbb{R}}
\newcommand{\E}{\mathbf{E}}

\title{Gradient-matching coresets for continual learning}

\author{%
  Lukas Balles, Giovanni Zappella, Cédric Archambeau \\
  Amazon Web Services, Berlin\\
  \texttt{\{balleslb, zappella, cedrica\}@amazon.de} \\
}

\begin{document}

\maketitle

\begin{abstract}
  We devise a coreset
  selection method based on the idea of gradient matching:
  the gradients induced by the coreset should match, as closely as possible, those
  induced by the original training dataset.
  We evaluate the method in the context of continual learning, where it
  can be used to curate a rehearsal memory.
  Our method performs strong competitors such as reservoir sampling
  across a range of memory sizes.
\end{abstract}

\section{Introduction}

Continual learning refers to the training of a machine learning model on
a sequence of data batches---referred to as \emph{tasks} in some scenarios---which
are sampled in a non-iid fashion.
Naive incremental training will lead to so-called
\emph{catastrophic forgetting}, where performance on previously observed data
deteriorates while training on new data.
Among the most successful strategies to counteract forgetting is the use of a
rehearsal memory of data points to be replayed while training on new
data.
In fact, \citet{prabhu2020gdumb} demonstrated that retraining the model from
scratch on an appropriately curated memory outperforms several
more complex methods.
Hence, maintaining an informative subset of a sequence of non-iid data batches
is a crucial task to enable continual learning.

This paper presents a coreset selection method based on the idea of \emph{gradient
matching.}
Consider a supervised learning task of predicting target $y\in\mathcal{Y}$ from
inputs $x\in \mathcal{X}$.
Assume we have a model parametrized by $\theta$.
A datapoint $(x, y)$ incurs a loss $\ell(\theta; x, y)$
and the mean loss induced by a training dataset $T\subset \mathcal{X}\times \mathcal{Y}$ is
$L_T(\theta) = \frac{1}{\vert T\vert} \sum_{(x, y) \in T} \ell(\theta; x, y)$.
The model interacts with the dataset through a series of evaluations of the
gradient $\nabla L_T(\theta)$.
Thus, if we can select a subset $C \subset T$
such that $\nabla L_C(\theta)
\approx \nabla L_T(\theta)$ for all
$\theta$, we can expect to preserve most of the information
that is relevant for training the chosen model.
Given a distribution $p(\theta)$ over the model parameters, we define the
gradient matching error of $C$ as follows:
\begin{equation}
	\label{eq:gradient-matching-criterion}
	\E_{p(\theta)} \left[
		\Vert \nabla L_C(\theta) - \nabla L_T(\theta) \Vert^2
	\right].
\end{equation}
This paper proposes an efficient method that uses this
criterion to select a weighted subset of the original dataset.
First, by constructing a finite-dimensional embedding of the gradient functions,
we reduce the problem to a cardinality-constrained quadratic optimization problem.
While NP-hard, it can be solved approximately using greedy methods.
Second, we show empirically that choosing $p(\theta)$ to be the
\emph{initialization distribution}
of the chosen (neural network) model is
already informative enough to extract good coresets.
Finally, we explain how the algorithm can be applied to the continual setting
and evaluate it experimentally.

\section{Related work}
\label{related}
A number of methods for efficientily managing memory in continual learning have been proposed
over time (e.g., see \cite{borsos2020coresets, aljundi2019online, aljundi2019gradient}). Due to space constraints we will
discuss only the ones which are closely related to our work.

\citet{zhao2020dataset} use a similar criterion as Eq.~\eqref{eq:gradient-matching-criterion}
to construct a \emph{synthetic} dataset using gradient-based
optimization w.r.t.~the (randomly initialized) data $C$.
Instead of using a distribution $p(\theta)$, they use points along training trajectories.
Their method achieves good results, but experiments are restricted to small coresets since the complexity
of optimizing $C$ grows with the allocated number of synthetic data points.
The method also exposes a number of sensitive optimization-related hyperparameters.

\citet{campbell2018bayesian} use a a similar technique
to select coresets for efficient Bayesian inference with Monte-Carlo methods.
The goal is to select a subset that yields a good approximation of the posterior.
They choose $p(\theta)$ to be a Laplace approximation
to the posterior, which requires minimizing the loss on the full dataset and is,
thus, inapplicable to settings like continual learning.

Recent independent work by \citet{killamsetty2021grad} proposes a similar technique
but considers a different setting.
Their goal is to reduce the computational cost of an offline (non-continual)
learning problem.
Instead of a distribution $p(\theta)$, they perform gradient matching locally at
the \emph{current} iterate.
After training on the resulting coreset for a small number of epochs, they
repeat the coreset construction from scratch, using the latest iterate.
Since the whole dataset has to be retained for the repeated coreset construction
their method is inapplicable to the continual learning setting.

\section{Gradient-matching coresets}

Sections~\ref{sec:optimal-coreset}--\ref{sec:embeddings} introduce the algorithm, agnostic to the choice of $p(\theta)$.
Section~\ref{sec:matching-at-init} justifies the use of the model's initialization
distribution.
Section~\ref{sec:streaming-mode} describes the application to continual learning.

\subsection{Optimal coreset and greedy selection}
\label{sec:optimal-coreset}

Assume $T$ contains $N$ datapoints and define $g_i(\theta) = \nabla \ell(\theta; x_i, y_i)$, $i\in [N]$,
and $g(\theta) = \sum_{i=1}^N g_i(\theta)$.
The weighted subset of cardinality $n< N$ with minimal gradient matching error
can be represented by a sparse vector $\lambda \in \R^N$ and may be found by
solving the following optimization problem:
\begin{equation}
	\label{eq:optimal_coreset}
	\textstyle \min_{\lambda \in \R^N} \mathbf{E}_{p(\theta)} \left[ \textstyle
		\left\Vert \sum_{i=1}^N \lambda_i g_i(\theta) - g(\theta) \right\Vert^2
	\right]
	\text{ s.t. } \Vert \lambda\Vert_0 \leq n,
\end{equation}
where $\Vert \lambda \Vert_0 = \vert \{ i \mid \lambda_i \neq 0 \} \vert$ is the $\ell_0$ pseudo-norm
which counts non-zero elements.
We can rewrite the objective as $\min_{\lambda \in \R^N} \Vert G\lambda - g\Vert_\mathcal{G}^2$,
where $\mathcal{G}$ is the Hilbert space on which the gradient functions live,
equipped with the inner product
$\langle g, \tilde{g} \rangle = \mathbf{E}_{p(\theta)}[g(\theta)^T\tilde{g}(\theta)]$,
and $G \colon \R^N \to \mathcal{G}$ is the linear operator which maps
$\lambda \mapsto \sum_i\lambda_i g_i$.
This is a cardinality-constrained quadratic program, which is known to be NP-hard.
Approximate solutions may be obtained with greedy algorithms, as we will discuss shortly.
Working with the infinite-dimensional objects $g_i$ and
taking expecatations over $\theta$ is intractable in practice.
Hence, we work with a finite-dimensional representation of the gradient functions
and solve a problem of the form
\begin{equation}
	\label{eq:omp-problem-finite}
	\textstyle \min_{\lambda \in \R^N} \Vert G\lambda - g\Vert_2^2 \text{ s.t. } \Vert \lambda\Vert_0 \leq n, \quad G=[g_1, \dotsc, g_N]\in \R^{D\times N}, \quad g\in \R^{D}.
\end{equation}
We defer the discussion of the finite-dimensional embedding to Section~\ref{sec:embeddings}
and first discuss how to solve Eq.~\eqref{eq:omp-problem-finite} given such an embedding.

To approximate solutions to problem \eqref{eq:omp-problem-finite} we use greedy selection with matching
pursuit \citep{mallat1993matching}.
Assume we currently have a coreset $I\subset [N]$ with corresponding weights
$\gamma \in \R^{\vert I \vert}$.
This gives us an approximation $g \approx G_I\gamma$, where $G_I$ is the restriction of $G$
to the index set $I$.
Matching pursuit greedily adds the element which best matches the residual
$r = g - G_I \gamma$.
We use the popular \emph{orthogonal matching pursuit (OMP)} variant; after each
greedy iteration, OMP optimally readjusts the weights of all coreset
elements, $\min_{\gamma} \Vert G_I \gamma - g\Vert^2$, resulting in
$\gamma = (G_I^TG_I)^{-1}G_I^Tg$.
We stop the coreset construction when the desired size is reached.
Algorithm~\ref{alg:omp} provides pseudo-code.

\subsection{Finite-dimensional gradient embeddings}
\label{sec:embeddings}

We obtain finite-dimensional embeddings of the gradient functions
by sampling $s$ draws from $p(\theta)$, for each of which we evaluate $g_i(\theta)$.
Since storing these gradients would quickly exceed the memory for large models,
we perform dimensionality reduction to a $d$-dimensional representation.
The final embedding is the concatenation of the $s$ draws with total dimension $D = s d$.
We pursued two different dimensionality reduction strategies:
(i) projection onto $d$ random $\{\pm 1\}$-valued vectors;
(ii) following recent related work \citep{Ash2020Deep,killamsetty2021grad},
we use the gradients w.r.t.~the model's final layer, which  can be computed
efficiently without performing a full backward
pass and is therefore more suitable for large models.
This will be referred to as the ``last layer'' variant in experiments below.

\subsection{Gradient matching at initialization}
\label{sec:matching-at-init}

We choose $p(\theta)$ to be the model's initialization distribution and
our experiments show that this yields an informative selection criterion.
While this might seem surprising, research on the \emph{neural tangent kernel}
\citep{jacot2018neural} has argued that the behavior of overparametrized neural networks is characterized by
its gradients at initialization.
This is also reflected in recent related work \citep{paul2021deep} using
the gradient norm, averaged over the initialization distribution, as an importance score
for data points.
From a practical standpoint, using the initialization distribution allows us to
select a coreset \emph{before} training.

It is also worth noting that OMP tends to choose data points $k_\star$ with small
$\langle g_{k_\star}, g_i\rangle$ for all $i\in I$ already in the coreset.
Therefore, we might also interpret the method as simply choosing \emph{diverse} subsets
with respect to a similarty function given by the inner product of gradients at
initialization.

\subsection{Application to continual learning}
\label{sec:streaming-mode}
For continual learning, we would like to employ gradient matching corsets (GMC) to a non-iid 
data
batches which need to be processed sequentially.
For each incoming batch, we compute the corresponding gradient embedding matrix
$G^{(t)} = [g^{(t)}_1, \dotsc, g^{(t)}_{N_t}] \in \R^{D\times N_t}$.
The goal is to maintain a coreset which, after each new batch, matches the aggregate gradient of all data points seen so far, which we 
define as follows:
\begin{equation}
  \label{eq:target-vector}
	\textstyle g^{(1:t)} := \sum_{s=1}^t \sum_{i=1}^{N_t} g^{(s)}_i.
\end{equation}
Let $C^{(t-1)}$ denote the gradient embedding matrix of the coreset after processing
tasks $1$ through $t-1$.
Upon receiving $G^{(t)}$, we first update the ``target vector'' according to
Eq.~\eqref{eq:target-vector}.
We then run OMP with target $g^{(1:t)}$ and dictionary $G = [C^{(t-1)}, G^{(t)}]$
and store the gradient embedding matrix of the resulting coreset.
Algorithm~\ref{alg:streaming} provides pseudo-code.

Compared to an ``offline'' setting where $G^{(1)}, \dotsc, G^{(t)}$ are accessible
simultaneously, we use the exact same target vector $g^{(1:t)}$ but a limited
dictionary $[C^{(t-1)}, G^{(t)}]$.
Since $C^{(t-1)}$ is a coreset representative of $G^{(1)}, \dotsc, G^{(t-1)}$,
we can expect the loss in performance to be small.
We emphasize that the algorithm is free to remove and/or reweight elements form
$C^{(t-1)}$.

\begin{minipage}[t]{.49\textwidth}
  \centering
  \begin{algorithm}[H]
    \caption{Orthogonal Matching Pursuit}
    \label{alg:omp}
  \begin{algorithmic}
    \State \hspace{-10pt}\textsc{OMP}($G = [g_1, \dotsc,  g_N]$, $g$, $n$)
      \State $I \gets ()$ \Comment{Coreset indices}
      \State $\gamma \gets ()$ \Comment{Coreset weights}
      \While{$\vert I \vert < n$}
        \State $k_\ast = \argmax_k \frac{\langle g_k, r\rangle}{\Vert g_k\Vert}$, \, $r = g - G_I\gamma$
        \State $I \gets I \cup (k_\ast)$
        \State $\gamma \gets (G_I^T G_I)^{-1}G_I^T g$
      \EndWhile
      \State \Return $I, \gamma$
  \end{algorithmic}
  \end{algorithm}
\end{minipage}%
\hfill
\begin{minipage}[t]{.49\textwidth}
  \centering
  \begin{algorithm}[H]
    \caption{Continual GMC}
    \label{alg:streaming}
  \begin{algorithmic}
    \State $g \gets 0$
    \State $C \gets ()$
    \While{receiving $G^{(t)} = [g^{(t)}_1, \dotsc, g^{(t)}_{N_t}]$}
      \State $g \gets g + \sum_i g^{(t)}_i$
      \State $G \gets [C, G^{(t)}]$
      \State $I, \gamma \gets$ \textsc{omp}$(G, g, n)$
      \State $C \gets G_I$
    \EndWhile
  \end{algorithmic}
  \end{algorithm}
\end{minipage}%

\section{Experiments}

We now evaluate our coreset method in the continual learning setting.
We use the simple but effective \textsc{gdumb} strategy proposed
by \citet{prabhu2020gdumb}.
Upon receiving a new batch of data, \textsc{gdumb} updates its rehearsal
memory, reinitializes the model and trains it from scratch using only
the data in memory.
\citet{prabhu2020gdumb} used a greedy class-balancing sampler, but the strategy
can likewise be used with any other subsampling or coreset method.
We use this paradigm because it has been shown to outperform many more involved
methods (e.g. EWC \citep{kirkpatrick2017overcoming} or LwF \citep{li2017learning})
and it isolates the effect of the memory curation strategy, which is our main
interest.

We present experiments training a simple MLP on tabular datasets, as well as a
CNN on \textsc{Fashion-MNIST} and ResNet-18 \citep{he2016deep} on \textsc{Cifar-10}.
For the tabular datasets, we simulate a task-free continual learning scenario
by sorting the data points according to the value of a single feature and splitting
into $10$ equally sized batches.
For \textsc{Fashion-MNIST} and \textsc{Cifar-10}, we use the popular ``class-incremental''
scenario, where the dataset is divided into discrete tasks, each consisting of
two classes.
Appendix~\ref{experimental-details} contains more details.

\begin{figure}
	\centering
	\includegraphics[width=.45\textwidth]{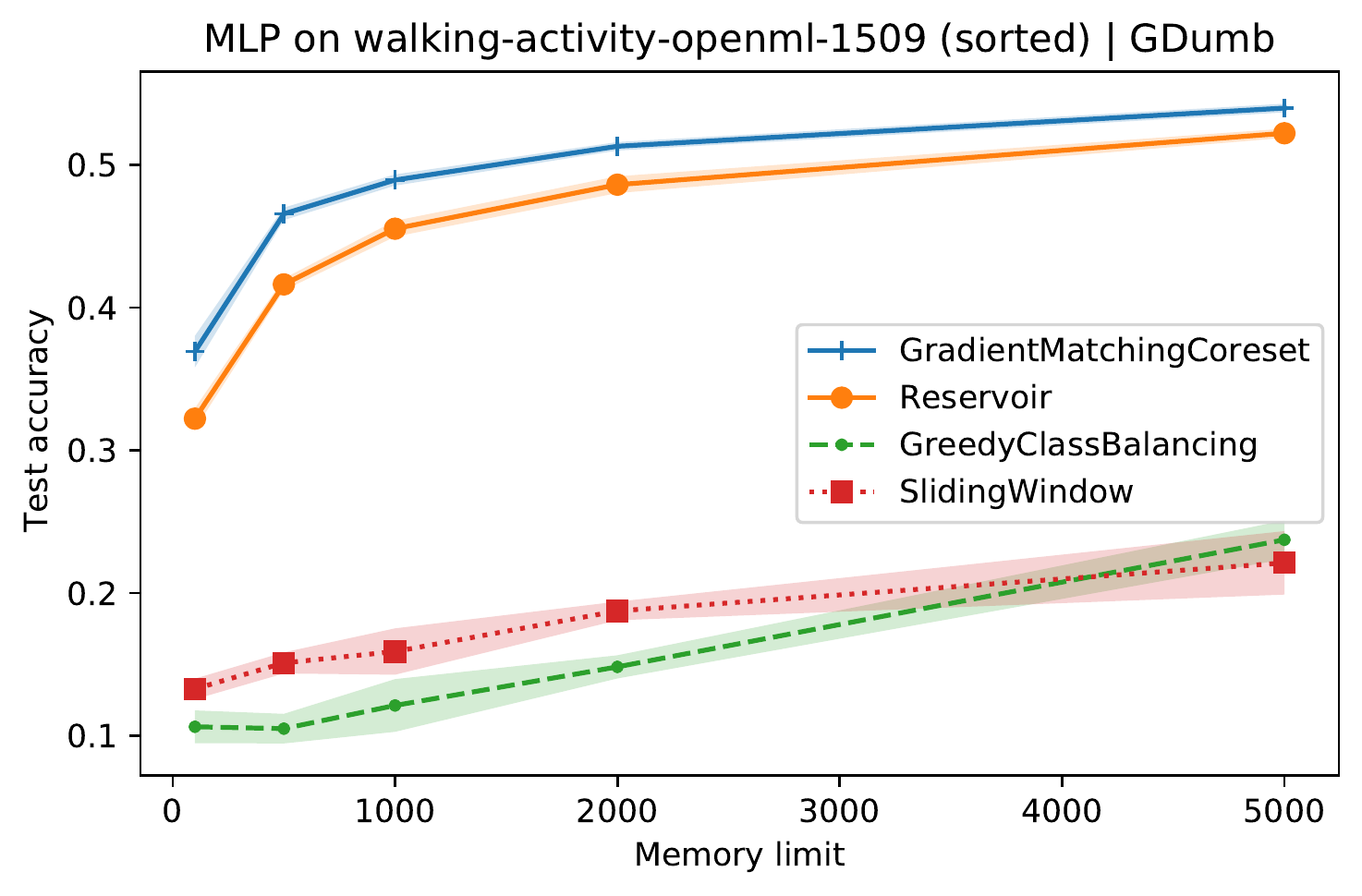}
	\includegraphics[width=.45\textwidth]{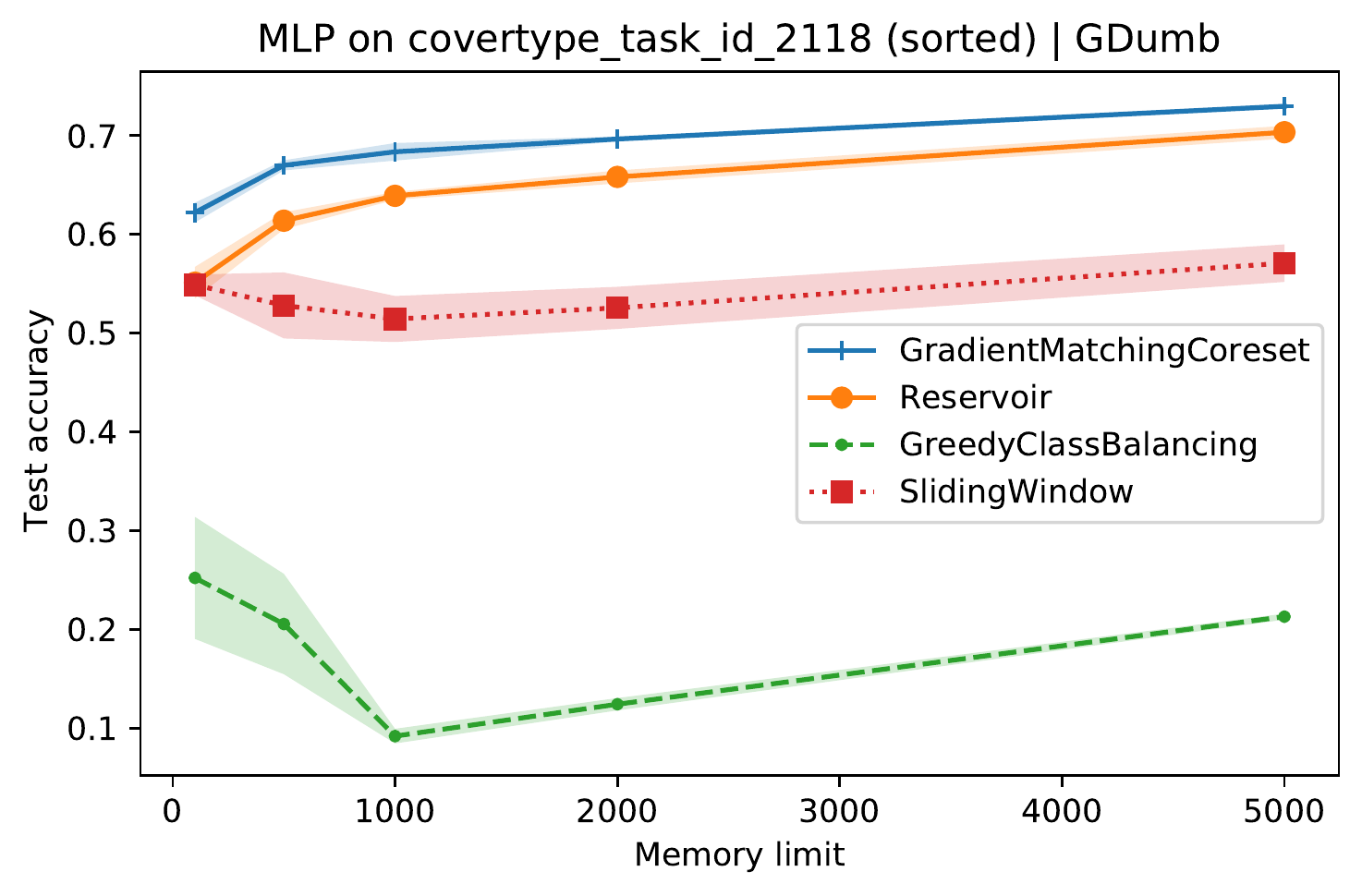}
  \includegraphics[width=.45\textwidth]{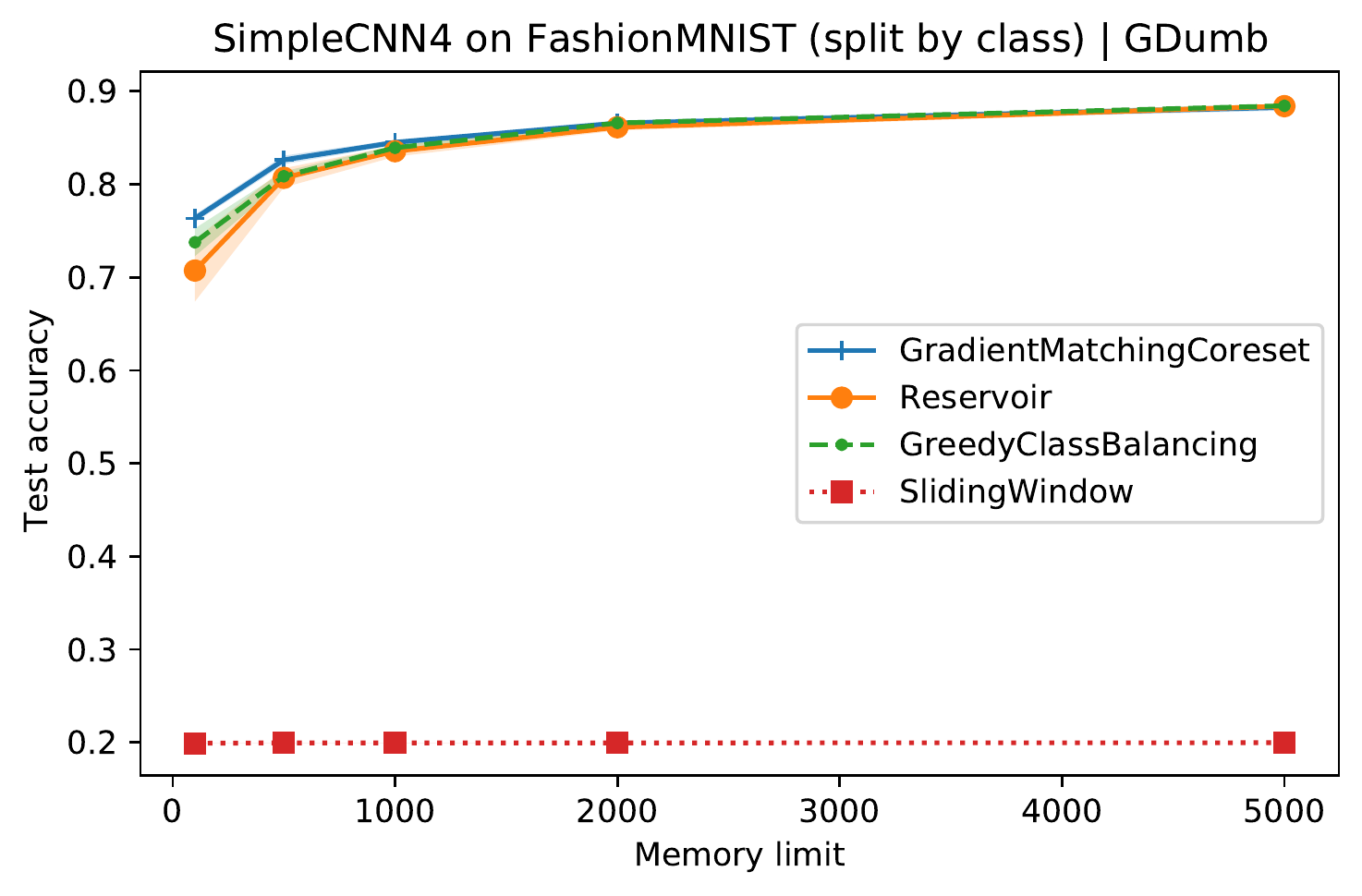}
	\includegraphics[width=.45\textwidth]{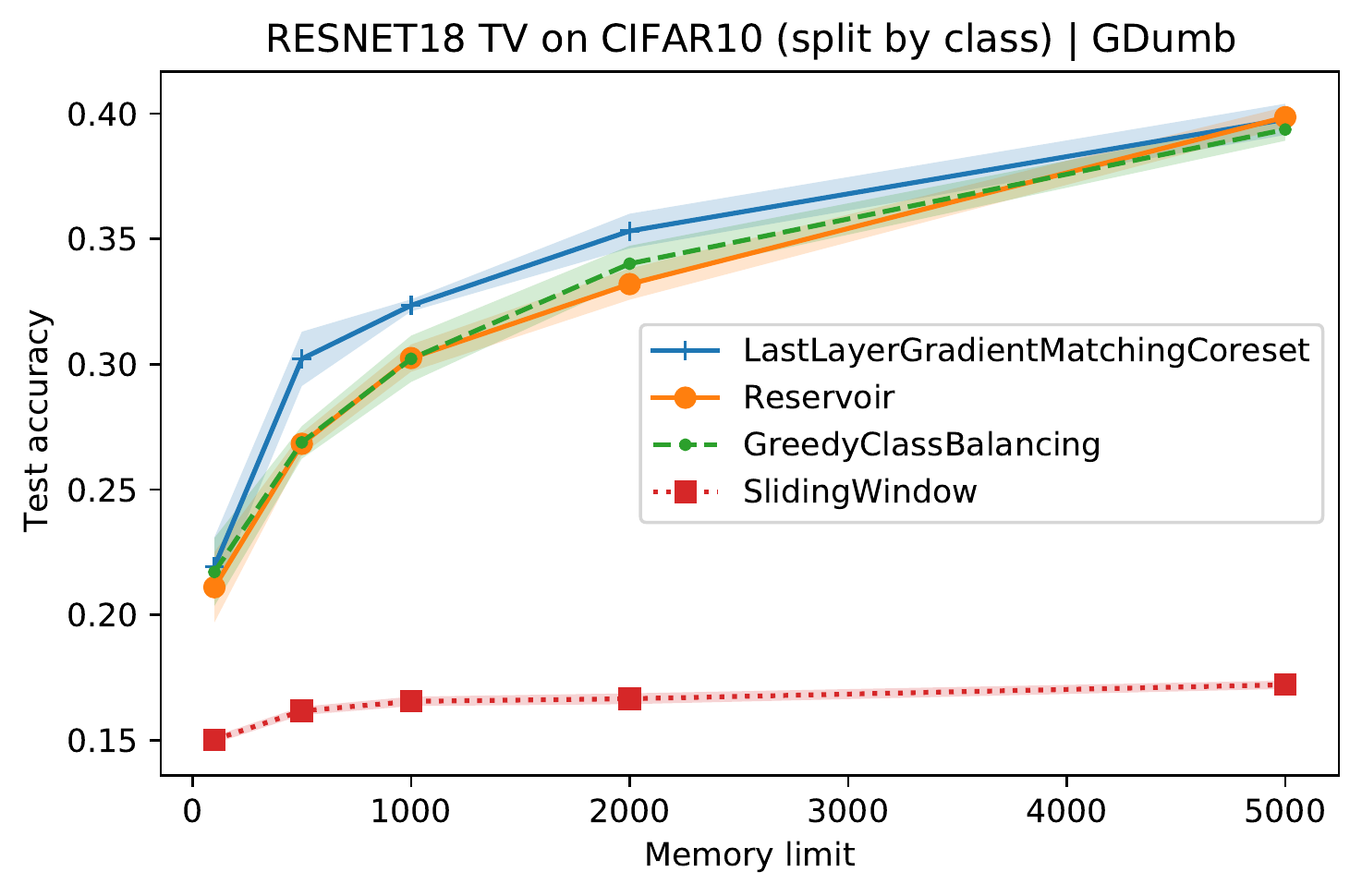}
	\caption{Results on continual learning scenarios using the \textsc{gdumb}
  paradigm with different subsampling/coreset methods. The graphs
  depict the final accuracy, after seeing all tasks/batches, on
  the full test set as a function of the memory size. Results are averaged over
  five random seeds and the shaded area spans one standard deviation.}
	\label{fig:results}
\end{figure}

We compare GMC to reservoir sampling \citet{vitter1985random}, which maintains
a uniform subsample of all data points seen so far, as well as the
greedy class-balancing method used by \citet{prabhu2020gdumb} and a naive
sliding window of the most recent data points.
We test memory sizes between $100$ and $5000$.
Figure~\ref{fig:results} depicts the results.
The sliding window heuristic fails due to the non-iid nature of the
continual learning scenarios.
The greedy class-balancing method performs well in the class-incremental scenario,
but fails in the task-free sorted scenario, where there's a continual structure
beyond a shift in class occurences.
We see that the gradient-matching coreset method achieves consistent improvements
over reservoir sampling at all tested memory sizes.
The relative improvement tends to be larger at small memory sizes.
More detailed plots with the performance over time in different experiments are
reported in Appendix~\ref{a:addresults}.
We also provide additional results using experience replay.

\section{Conclusion}

We  demonstrated that GMC is an effective method for coreset selection, applicable
to non-iid sequence of data batches.
It is simple, robust and scales to coreset sizes of several thousand data points.
Moreover, GMC does not expose any critical hyperparameters.
In the future, we would like to futher increase the scalability of the method to create larger coreset which can be useful
to train large neural networks from scratch.

\bibliographystyle{plainnat}
\bibliography{references.bib}

\begin{thebibliography}{19}
\providecommand{\natexlab}[1]{#1}
\providecommand{\url}[1]{\texttt{#1}}
\expandafter\ifx\csname urlstyle\endcsname\relax
  \providecommand{\doi}[1]{doi: #1}\else
  \providecommand{\doi}{doi: \begingroup \urlstyle{rm}\Url}\fi

\bibitem[Aljundi et~al.(2019{\natexlab{a}})Aljundi, Caccia, Belilovsky, Caccia,
  Lin, Charlin, and Tuytelaars]{aljundi2019online}
Rahaf Aljundi, Lucas Caccia, Eugene Belilovsky, Massimo Caccia, Min Lin,
  Laurent Charlin, and Tinne Tuytelaars.
\newblock Online continual learning with maximally interfered retrieval,
  2019{\natexlab{a}}.

\bibitem[Aljundi et~al.(2019{\natexlab{b}})Aljundi, Lin, Goujaud, and
  Bengio]{aljundi2019gradient}
Rahaf Aljundi, Min Lin, Baptiste Goujaud, and Yoshua Bengio.
\newblock Gradient based sample selection for online continual learning,
  2019{\natexlab{b}}.

\bibitem[Ash et~al.(2020)Ash, Zhang, Krishnamurthy, Langford, and
  Agarwal]{Ash2020Deep}
Jordan~T. Ash, Chicheng Zhang, Akshay Krishnamurthy, John Langford, and Alekh
  Agarwal.
\newblock Deep batch active learning by diverse, uncertain gradient lower
  bounds.
\newblock In \emph{International Conference on Learning Representations}, 2020.

\bibitem[Badanidiyuru et~al.(2014)Badanidiyuru, Mirzasoleiman, Karbasi, and
  Krause]{badanidiyuru2014streaming}
Ashwinkumar Badanidiyuru, Baharan Mirzasoleiman, Amin Karbasi, and Andreas
  Krause.
\newblock Streaming submodular maximization: Massive data summarization on the
  fly.
\newblock In \emph{Proceedings of the 20th ACM SIGKDD international conference
  on Knowledge discovery and data mining}, pages 671--680, 2014.

\bibitem[Borsos et~al.(2020)Borsos, Mutn{\`y}, and Krause]{borsos2020coresets}
Zal{\'a}n Borsos, Mojm{\'\i}r Mutn{\`y}, and Andreas Krause.
\newblock Coresets via bilevel optimization for continual learning and
  streaming.
\newblock \emph{arXiv preprint arXiv:2006.03875}, 2020.

\bibitem[Campbell and Broderick(2018)]{campbell2018bayesian}
Trevor Campbell and Tamara Broderick.
\newblock Bayesian coreset construction via greedy iterative geodesic ascent.
\newblock In \emph{International Conference on Machine Learning}, pages
  698--706. PMLR, 2018.

\bibitem[Chaudhry et~al.(2019)Chaudhry, Rohrbach, Elhoseiny, Ajanthan, Dokania,
  Torr, and Ranzato]{chaudhry2019tiny}
Arslan Chaudhry, Marcus Rohrbach, Mohamed Elhoseiny, Thalaiyasingam Ajanthan,
  Puneet~K Dokania, Philip~HS Torr, and Marc'Aurelio Ranzato.
\newblock On tiny episodic memories in continual learning.
\newblock \emph{arXiv preprint arXiv:1902.10486}, 2019.

\bibitem[He et~al.(2016)He, Zhang, Ren, and Sun]{he2016deep}
Kaiming He, Xiangyu Zhang, Shaoqing Ren, and Jian Sun.
\newblock Deep residual learning for image recognition.
\newblock In \emph{Proceedings of the IEEE conference on computer vision and
  pattern recognition}, pages 770--778, 2016.

\bibitem[Jacot et~al.(2018)Jacot, Gabriel, and Hongler]{jacot2018neural}
Arthur Jacot, Franck Gabriel, and Clement Hongler.
\newblock Neural tangent kernel: Convergence and generalization in neural
  networks.
\newblock In S.~Bengio, H.~Wallach, H.~Larochelle, K.~Grauman, N.~Cesa-Bianchi,
  and R.~Garnett, editors, \emph{Advances in Neural Information Processing
  Systems}, volume~31. Curran Associates, Inc., 2018.
\newblock URL
  \url{https://proceedings.neurips.cc/paper/2018/file/5a4be1fa34e62bb8a6ec6b91d2462f5a-Paper.pdf}.

\bibitem[Killamsetty et~al.(2021)Killamsetty, Sivasubramanian, Mirzasoleiman,
  Ramakrishnan, De, and Iyer]{killamsetty2021grad}
Krishnateja Killamsetty, Durga Sivasubramanian, Baharan Mirzasoleiman, Ganesh
  Ramakrishnan, Abir De, and Rishabh Iyer.
\newblock Grad-match: A gradient matching based data subset selection for
  efficient learning.
\newblock \emph{arXiv preprint arXiv:2103.00123}, 2021.

\bibitem[Kirkpatrick et~al.(2017)Kirkpatrick, Pascanu, Rabinowitz, Veness,
  Desjardins, Rusu, Milan, Quan, Ramalho, Grabska-Barwinska,
  et~al.]{kirkpatrick2017overcoming}
James Kirkpatrick, Razvan Pascanu, Neil Rabinowitz, Joel Veness, Guillaume
  Desjardins, Andrei~A Rusu, Kieran Milan, John Quan, Tiago Ramalho, Agnieszka
  Grabska-Barwinska, et~al.
\newblock Overcoming catastrophic forgetting in neural networks.
\newblock \emph{Proceedings of the national academy of sciences}, 114\penalty0
  (13):\penalty0 3521--3526, 2017.

\bibitem[Li and Hoiem(2017)]{li2017learning}
Zhizhong Li and Derek Hoiem.
\newblock Learning without forgetting.
\newblock \emph{IEEE transactions on pattern analysis and machine
  intelligence}, 40\penalty0 (12):\penalty0 2935--2947, 2017.

\bibitem[Mallat and Zhang(1993)]{mallat1993matching}
St{\'e}phane~G Mallat and Zhifeng Zhang.
\newblock Matching pursuits with time-frequency dictionaries.
\newblock \emph{IEEE Transactions on signal processing}, 41\penalty0
  (12):\penalty0 3397--3415, 1993.

\bibitem[Paszke et~al.(2019)Paszke, Gross, Massa, Lerer, Bradbury, Chanan,
  Killeen, Lin, Gimelshein, Antiga, Desmaison, Kopf, Yang, DeVito, Raison,
  Tejani, Chilamkurthy, Steiner, Fang, Bai, and Chintala]{paszke2019pytorch}
Adam Paszke, Sam Gross, Francisco Massa, Adam Lerer, James Bradbury, Gregory
  Chanan, Trevor Killeen, Zeming Lin, Natalia Gimelshein, Luca Antiga, Alban
  Desmaison, Andreas Kopf, Edward Yang, Zachary DeVito, Martin Raison, Alykhan
  Tejani, Sasank Chilamkurthy, Benoit Steiner, Lu~Fang, Junjie Bai, and Soumith
  Chintala.
\newblock Pytorch: An imperative style, high-performance deep learning library.
\newblock In H.~Wallach, H.~Larochelle, A.~Beygelzimer, F.~d\textquotesingle
  Alch\'{e}-Buc, E.~Fox, and R.~Garnett, editors, \emph{Advances in Neural
  Information Processing Systems 32}, pages 8024--8035. Curran Associates,
  Inc., 2019.
\newblock URL
  \url{http://papers.neurips.cc/paper/9015-pytorch-an-imperative-style-high-performance-deep-learning-library.pdf}.

\bibitem[Paul et~al.(2021)Paul, Ganguli, and Dziugaite]{paul2021deep}
Mansheej Paul, Surya Ganguli, and Gintare~Karolina Dziugaite.
\newblock Deep learning on a data diet: Finding important examples early in
  training.
\newblock \emph{arXiv preprint arXiv:2107.07075}, 2021.

\bibitem[Prabhu et~al.(2020)Prabhu, Torr, and Dokania]{prabhu2020gdumb}
Ameya Prabhu, Philip~HS Torr, and Puneet~K Dokania.
\newblock Gdumb: A simple approach that questions our progress in continual
  learning.
\newblock In \emph{European conference on computer vision}, pages 524--540.
  Springer, 2020.

\bibitem[Rubinstein et~al.(2008)Rubinstein, Zibulevsky, and
  Elad]{rubinstein2008efficient}
Ron Rubinstein, Michael Zibulevsky, and Michael Elad.
\newblock Efficient implementation of the k-svd algorithm using batch
  orthogonal matching pursuit.
\newblock Technical report, Computer Science Department, Technion, 2008.

\bibitem[Vitter(1985)]{vitter1985random}
Jeffrey~S Vitter.
\newblock Random sampling with a reservoir.
\newblock \emph{ACM Transactions on Mathematical Software (TOMS)}, 11\penalty0
  (1):\penalty0 37--57, 1985.

\bibitem[Zhao et~al.(2020)Zhao, Mopuri, and Bilen]{zhao2020dataset}
Bo~Zhao, Konda~Reddy Mopuri, and Hakan Bilen.
\newblock Dataset condensation with gradient matching.
\newblock \emph{arXiv preprint arXiv:2006.05929}, 2020.

\end{thebibliography}

\newpage
\appendix

\section{Additional results}
\label{a:addresults}
\subsection{Continual performance}

\begin{figure}
  \centering
  \includegraphics[width=.49\textwidth]{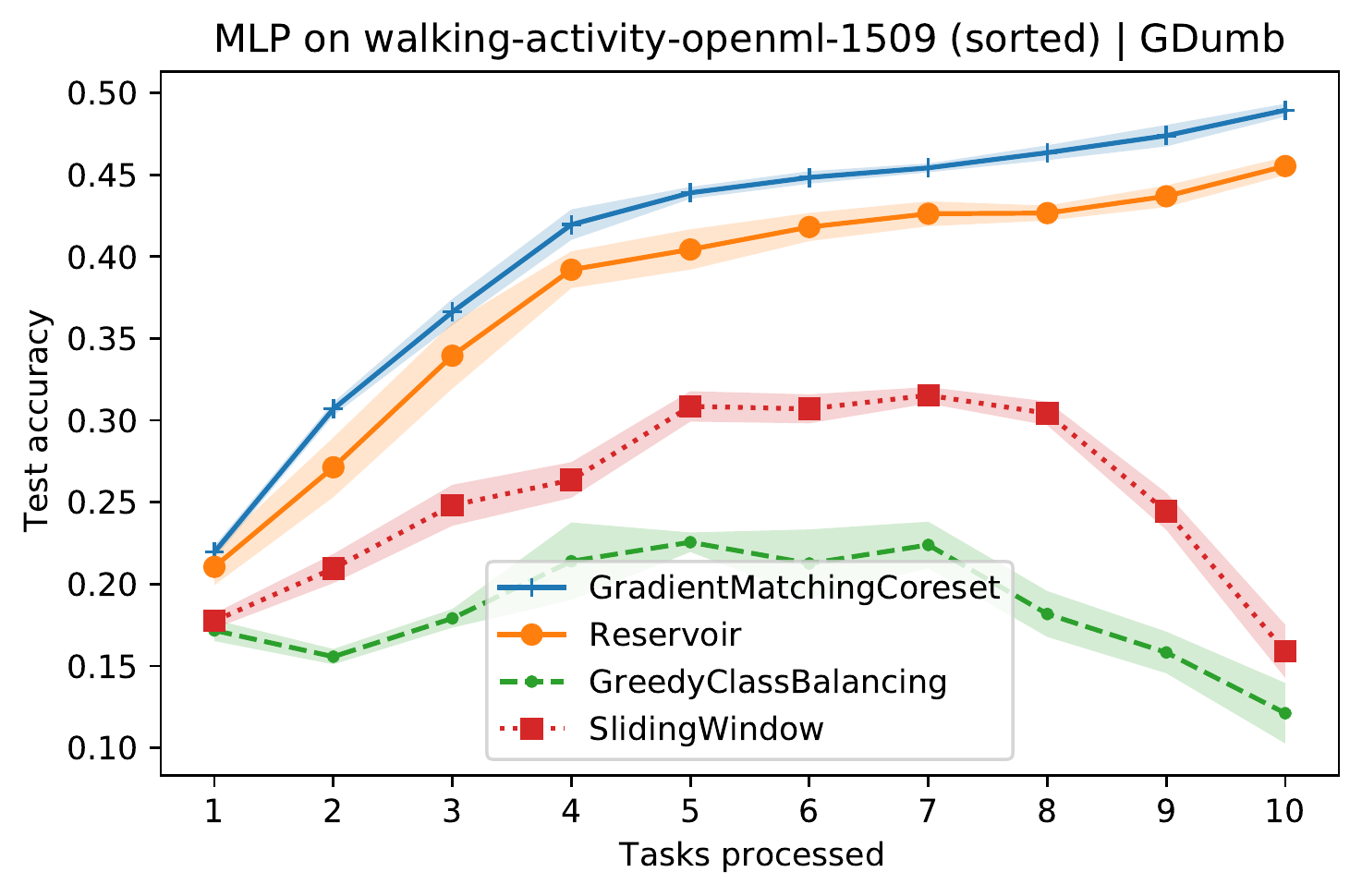}
  \includegraphics[width=.49\textwidth]{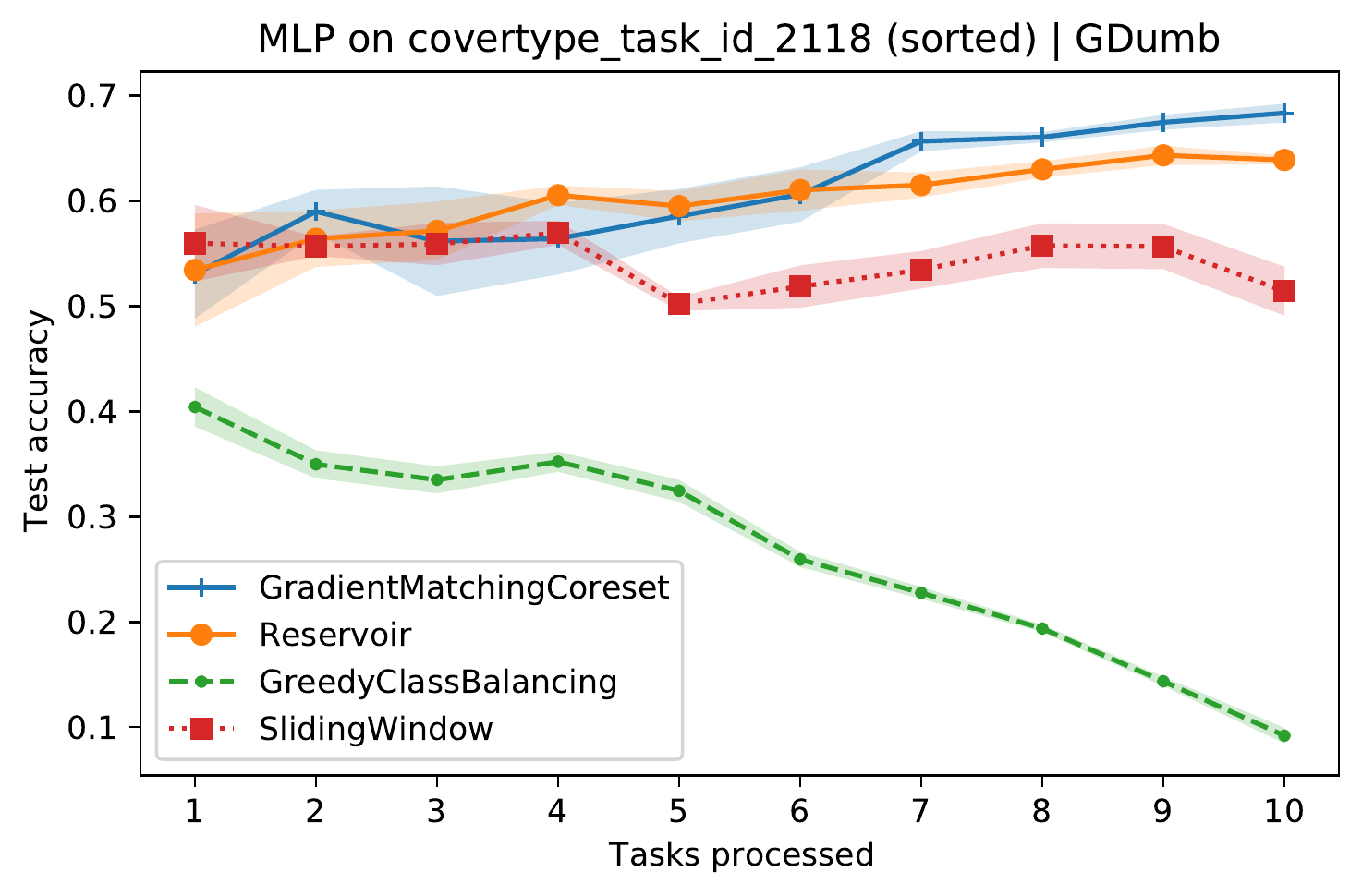}
  \includegraphics[width=.49\textwidth]{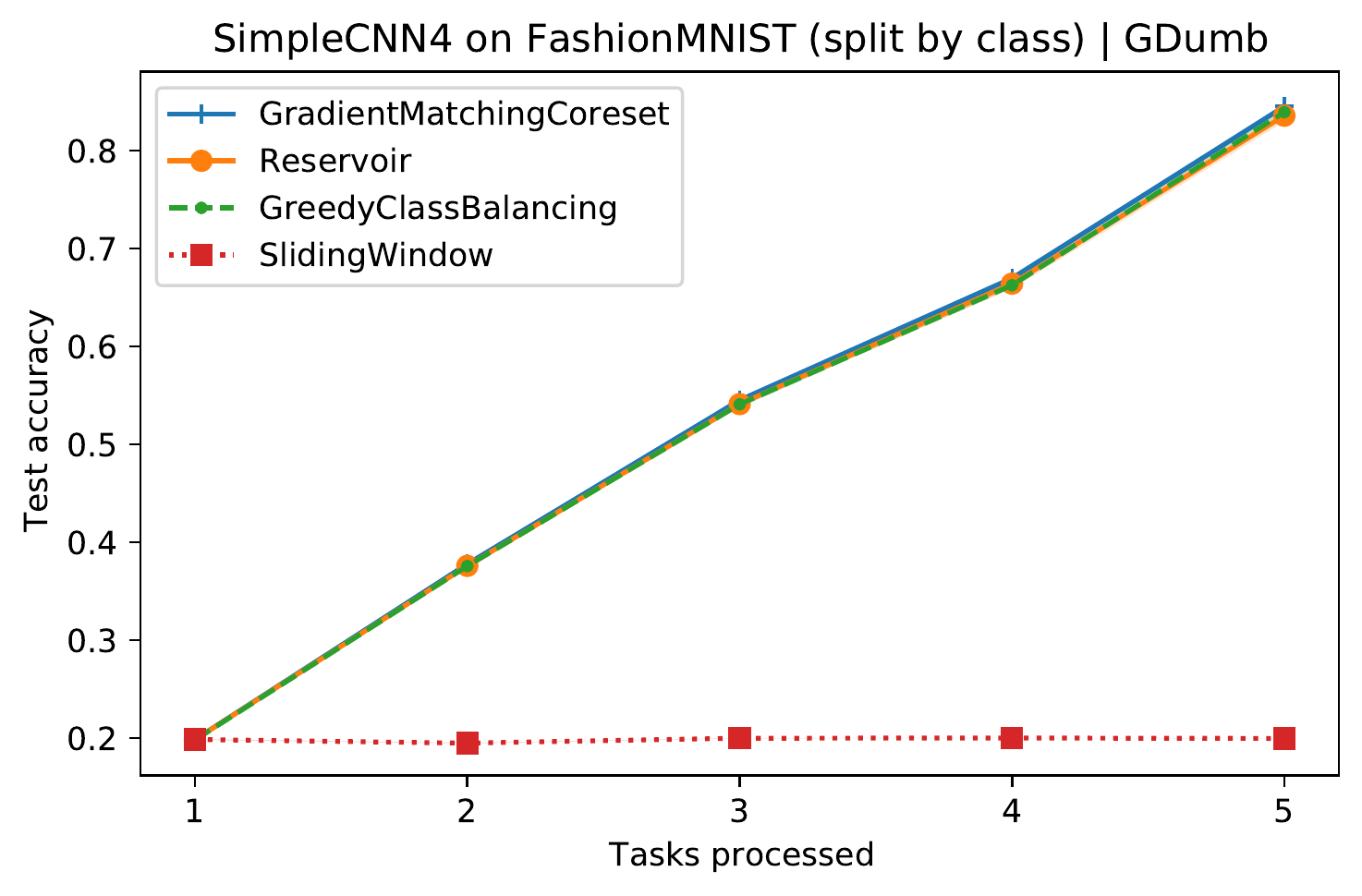}
  \includegraphics[width=.49\textwidth]{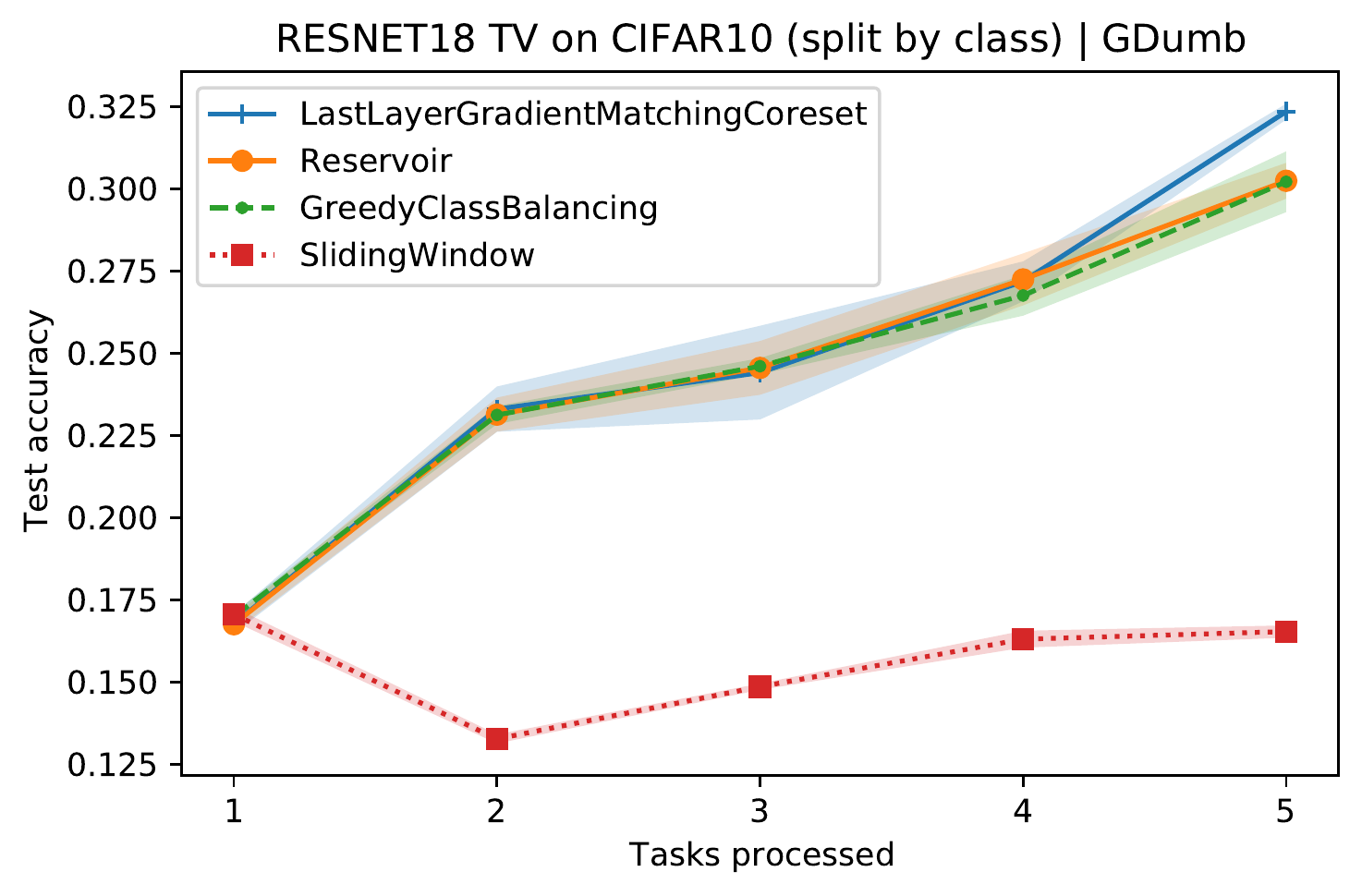}
  \caption{Results on continual learning scenarios using the \textsc{gdumb}
  paradigm with different subsampling/coreset methods at a memory size of $1000$.
  The graphs depict the accuracy on
  the full test set after the processing of each task/batch. Results are averaged over
  five random seeds and the shaded area spans one standard deviation.}
  \label{fig:gdumb-results-per-task}
\end{figure}

The plots in the main text depict the performance after processing all tasks.
To get a more fine-grained few of the continual behavior,
Figure~\ref{fig:gdumb-results-per-task} depicts the performance after each
individual task.
It shows results from the same experiment that underlies Figure~\ref{fig:results}
but only shows a single memory size of $1000$ for readability.

\subsection{Facility location baseline}

\begin{figure}
  \includegraphics[width=.49\textwidth]{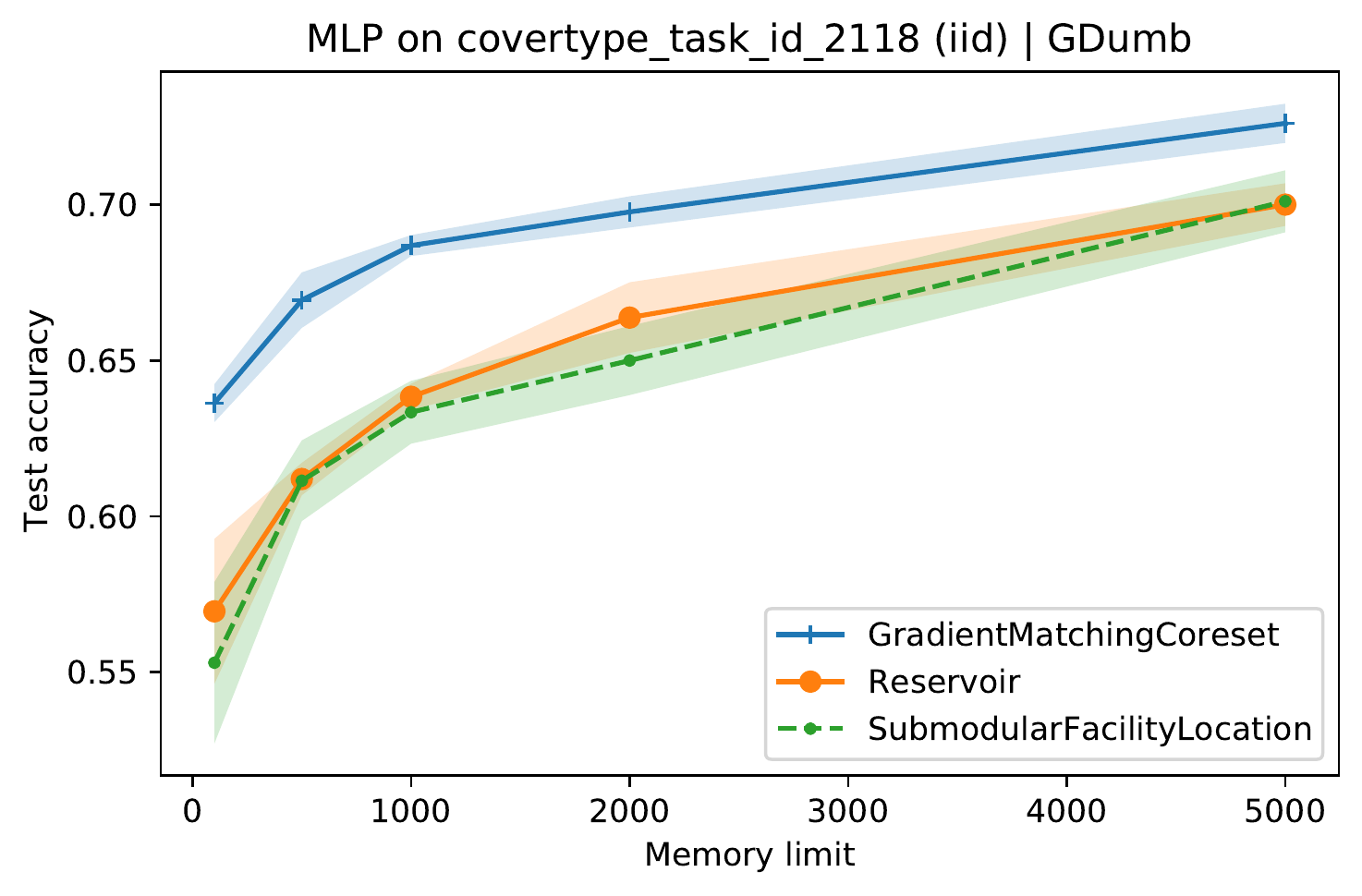}
  \includegraphics[width=.49\textwidth]{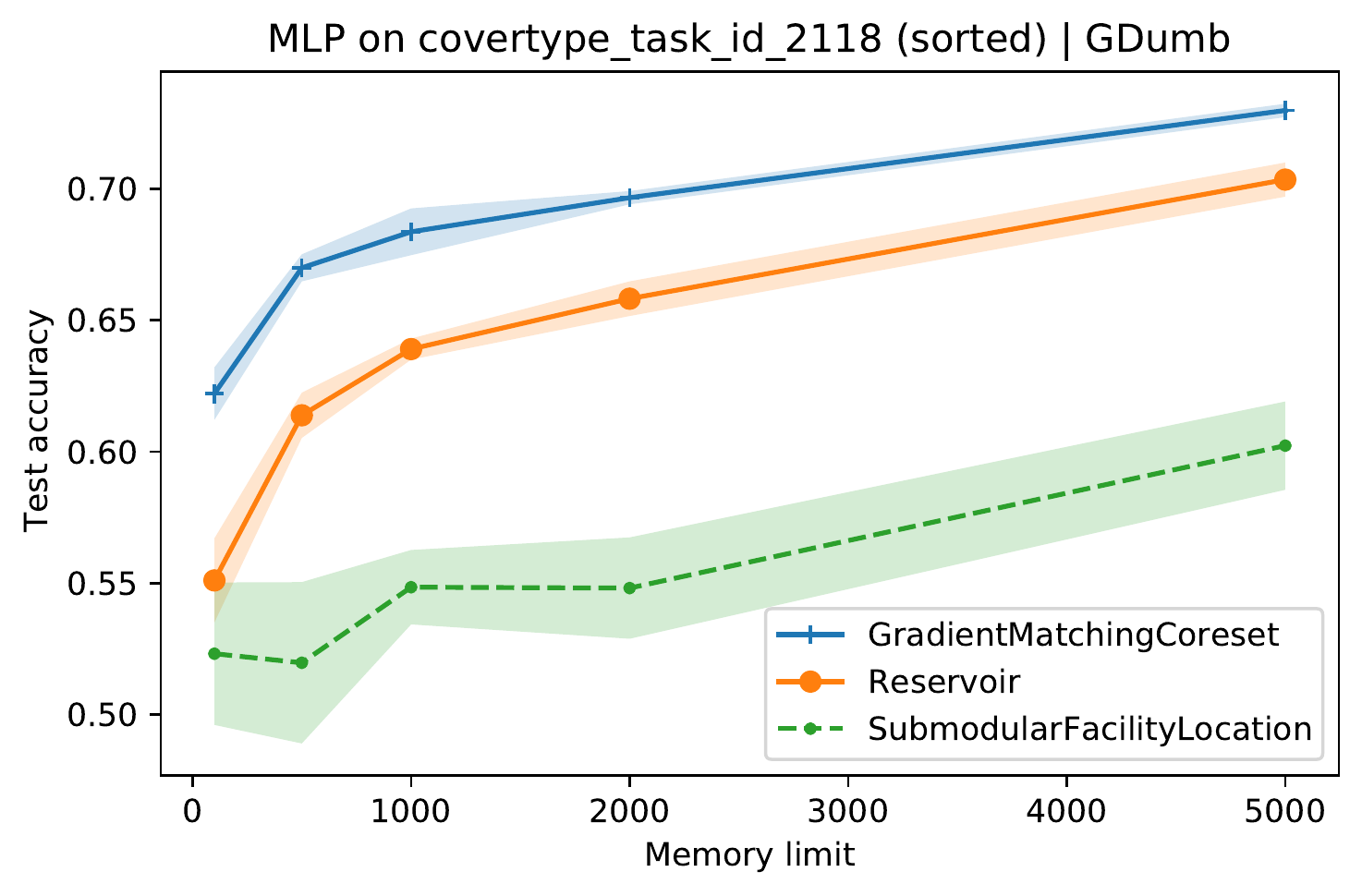}
  \caption{Results for the submodular facility location method using the \textsc{gdumb}
  paradigm.
  Note that we display results on a single dataset, but two different continual
  scenarios: iid-incremental and sorted.
  The graphs depict the final accuracy, after seeing all tasks/batches, on
  the full test set as a function of the memory size. Results are averaged over
  five random seeds and the shaded area spans one standard deviation.
  The facility location method matches, but fails to outperform, reservoir sampling in the iid-incremental
  setting.
  In the sorted setting, the method fails.}
  \label{fig:results-submod}
\end{figure}

As a baseline for our gradient-matching coreset method, we experimented with a
coreset based on the solution of a facility location problem in feature space,
i.e., choose $C\subset T$ such as to minimize
\begin{equation}
  \sum_{x\in T} \max_{x^\prime \in C} \Vert x - x^\prime\Vert.
\end{equation}
This is a submodular function and can be optimized approximately with greedy
submodular optimization methods.
In the continual setting, this has to be done in a streaming fashion; we followed
the \emph{sieve streaming} approach of \citet{badanidiyuru2014streaming} and
relied on an implementation in the open source Python package \texttt{apricot-select}.
We depict results on Covertype in Fig.~\ref{fig:results-submod}.
We show two different continual learning scenarios; our usual sorted scenario and,
as a sanity check, a simple ``iid-incremental'' scenario, where batches consist
of uniform subsamples of the dataset.
While the facility location method matches, but fails to outperform, reservoir sampling in the iid-incremental
setting, it fails in the sorted setting.

\subsection{Experience replay}

\begin{figure}
	\centering
  \includegraphics[width=.49\textwidth]{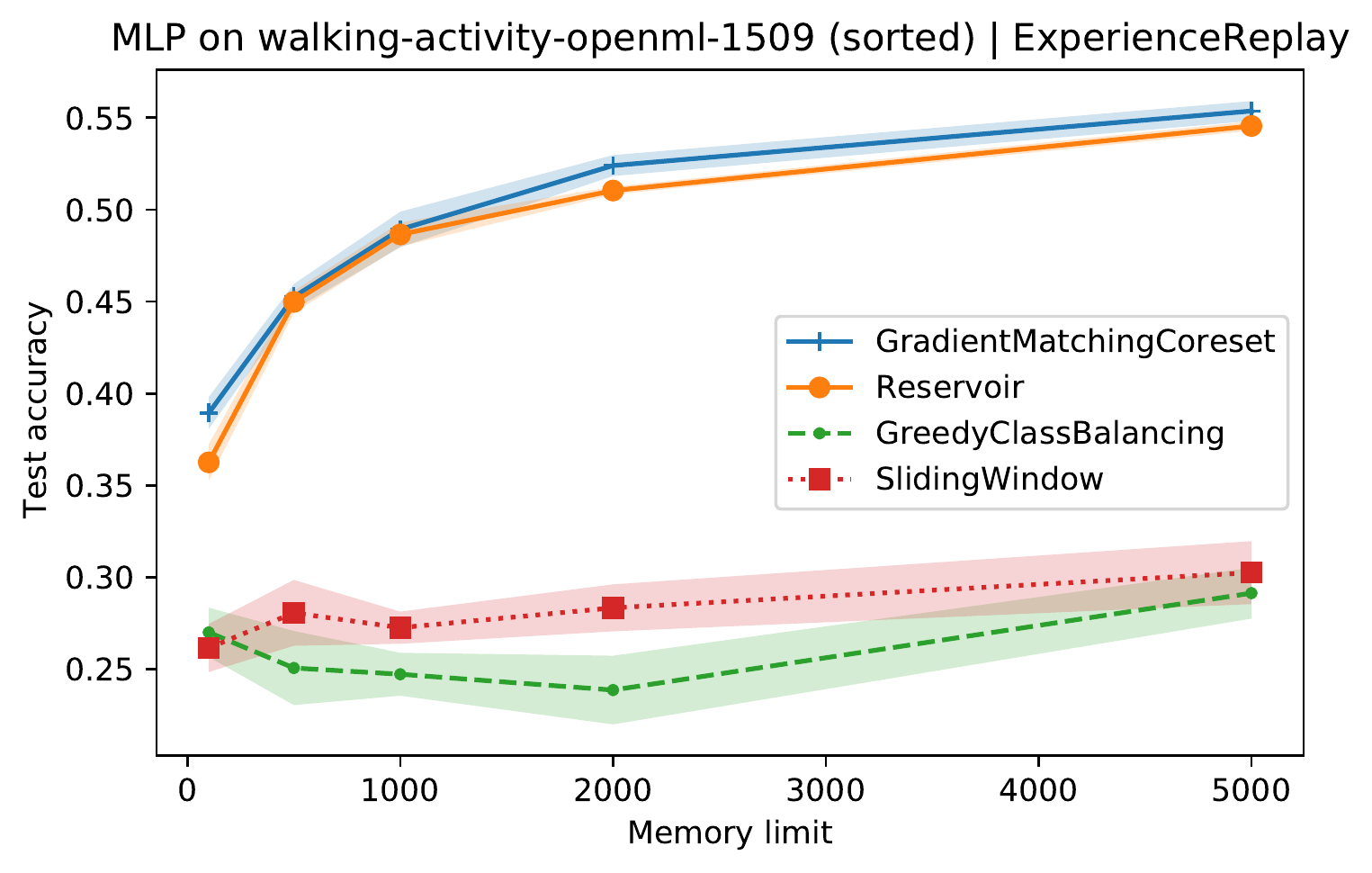}
	\includegraphics[width=.49\textwidth]{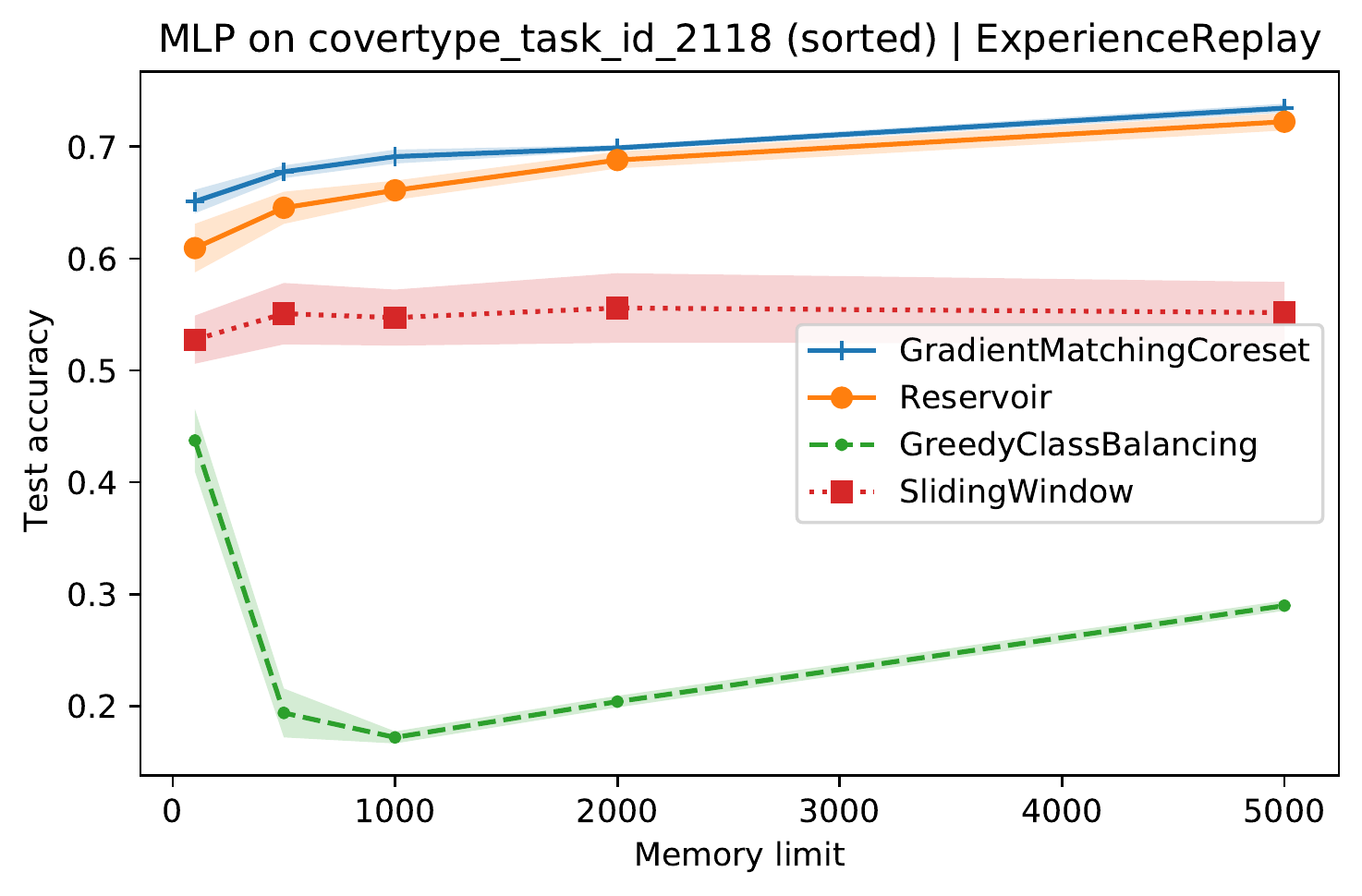}
  \includegraphics[width=.49\textwidth]{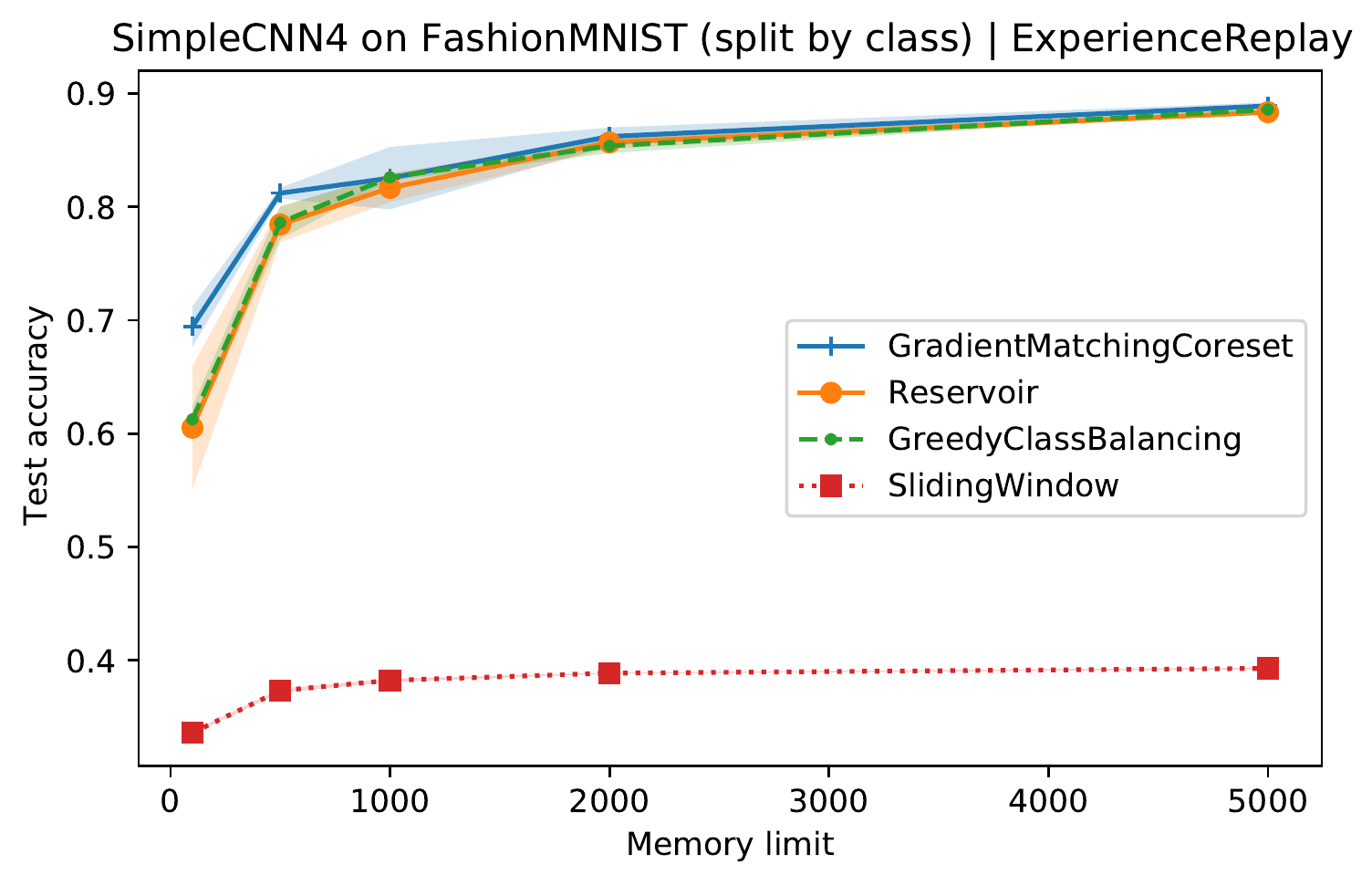}
  \includegraphics[width=.49\textwidth]{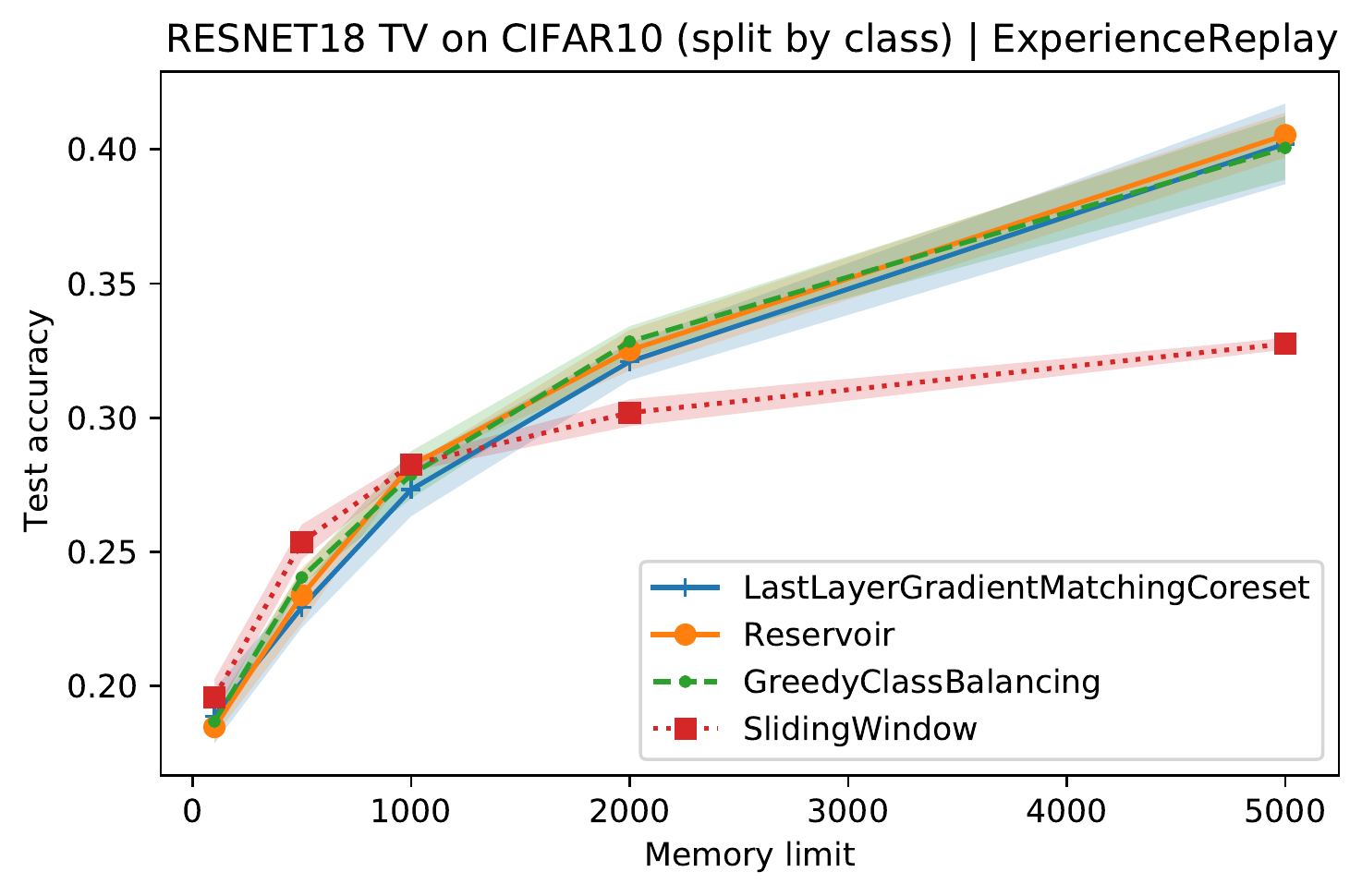}
	\caption{Results on continual learning scenarios using the experience replay
  paradigm with different subsampling/coreset methods. The graphs
  depict the final accuracy, after seeing all tasks/batches, on
  the full test set as a function of the memory size. Results are averaged over
  five random seeds and the shaded area spans one standard deviation.}
	\label{fig:er-results}
\end{figure}

Figure~\ref{fig:er-results} shows results using the experience replay method.
In contrast to \textsc{gdumb}, experience replay does not reinitialize the model
after receiving a new batch of data. Instead, it resumes training from the
previously found solution and trains on an (appropriately weighted) combination
of the current batch and the stored memory (see, e.g., \citet{chaudhry2019tiny}).
The memory is updated after the processing of each task.
Like with \textsc{gdumb}, our coreset method outperforms reservoir
sampling, albeit by a smaller margin.
In the \textsc{cifar-10} experiment, all tested subsampling/coreset methods
have almost identical performance.

\subsection{Local variant of GMC}

\begin{figure}
  \centering
  \includegraphics[width=.49\textwidth]{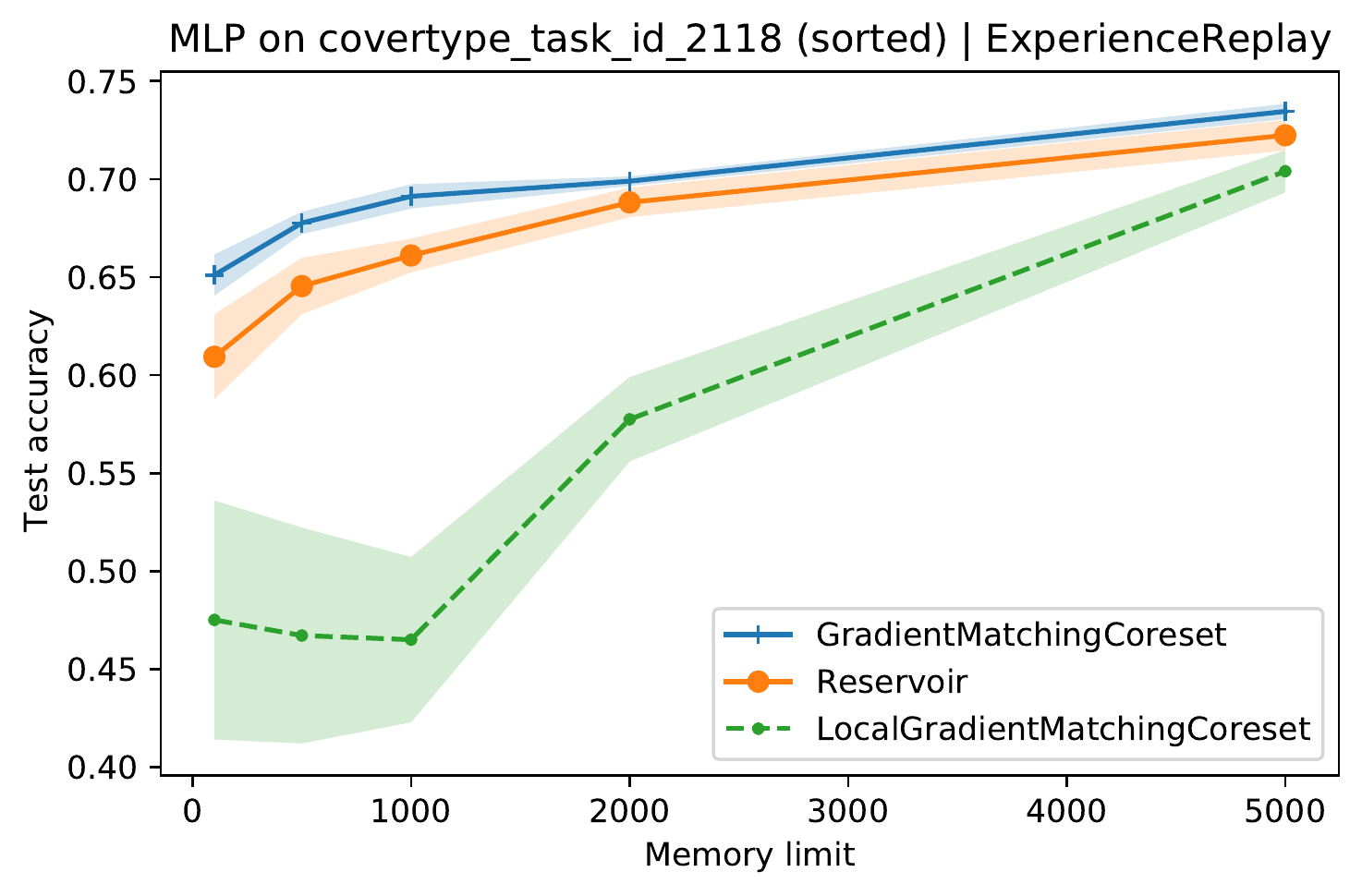}
  \includegraphics[width=.49\textwidth]{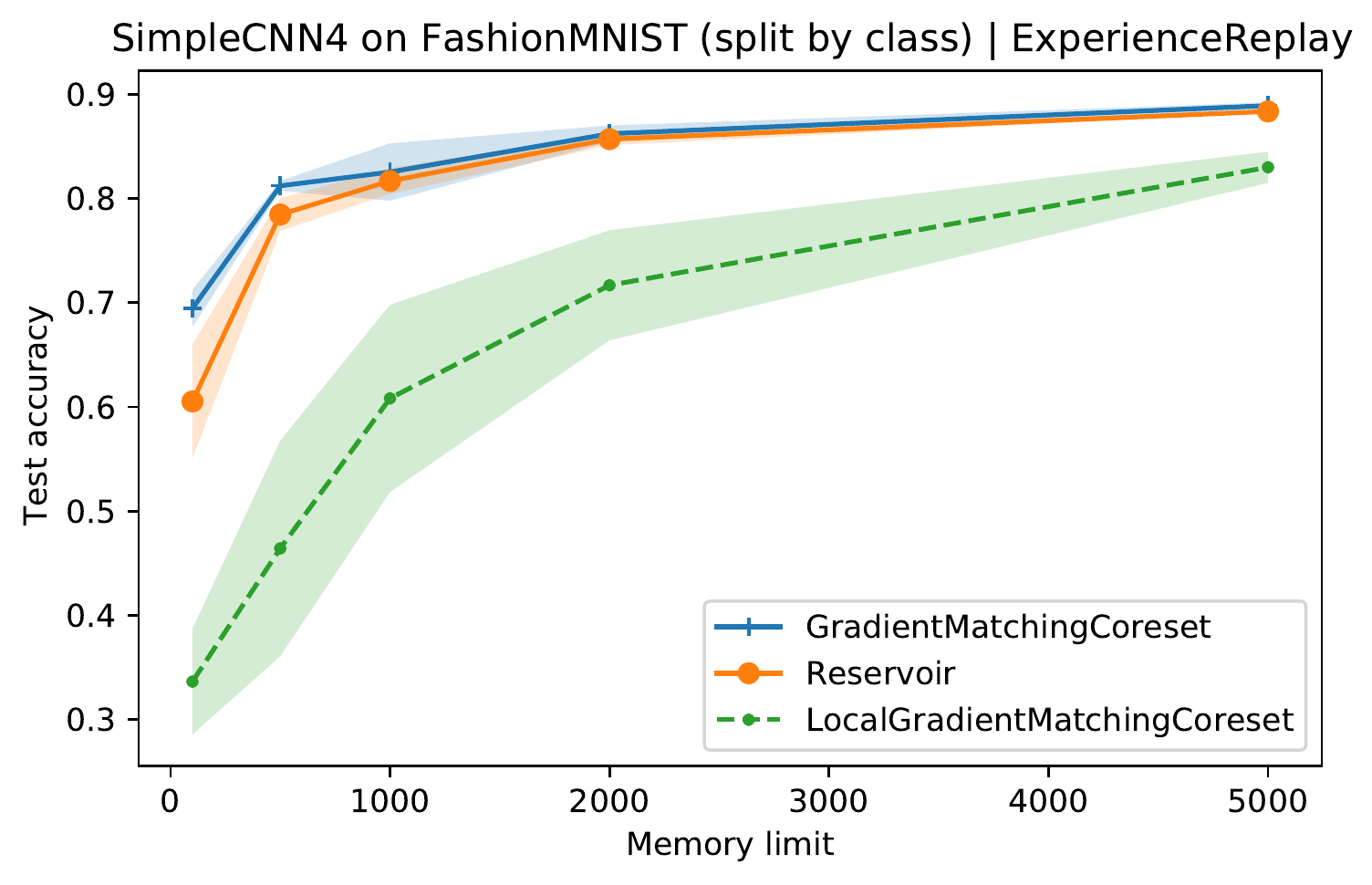}
  \caption{Results of the local variant of GMC on continual learning scenarios using the experience replay
  paradigm with different subsampling/coreset methods. The graphs
  depict the final accuracy, after seeing all tasks/batches, on
  the full test set as a function of the memory size. Results are averaged over
  five random seeds and the shaded area spans one standard deviation.}
  \label{fig:local-variant}
\end{figure}

We also performed initial exploratory experiments with a ``local'' variant of
GMC, designed to work with experience replay.
After training on a task and the current memory, we perform gradient
matching locally at the latest iterate and replace the memory with the obtained
coreset.
Since the gradients used for gradient matching now change over time, this requires
a small change in Continual GMC (Alg.~\ref{alg:streaming}).
Instead of storing the gradient embedding matrix of the coreset for use in the
next iteration, we need to recompute it in each iteration.
The resulting method can be seen as an adaptation of the method proposed by
\citet{killamsetty2021grad} to the continual scenario.
Unfortunately, the results, depicted in Fig.~\ref{fig:local-variant}, are much
worse.
We conjecture that the gradients at a single point
in weight space contain too little information to select coresets that are useful
beyond a small number of training epochs.

\section{Experimental details}
\label{experimental-details}

\subsection{Continual learning scenarios}

As mentioned in the main text, the class-incremental scenario on \textsc{FashionMNIST}
and \textsc{Cifar-10} is obtained by splitting the $10$ classes of the dataset
into $5$ tasks, consisting of classes $\{0, 1\}, \{2, 3\}, \dotsc, \{8, 9\}$.
The evaluation is done on the entire test set containing all classes.

\begin{figure}
  \centering
  \includegraphics[width=.49\textwidth]{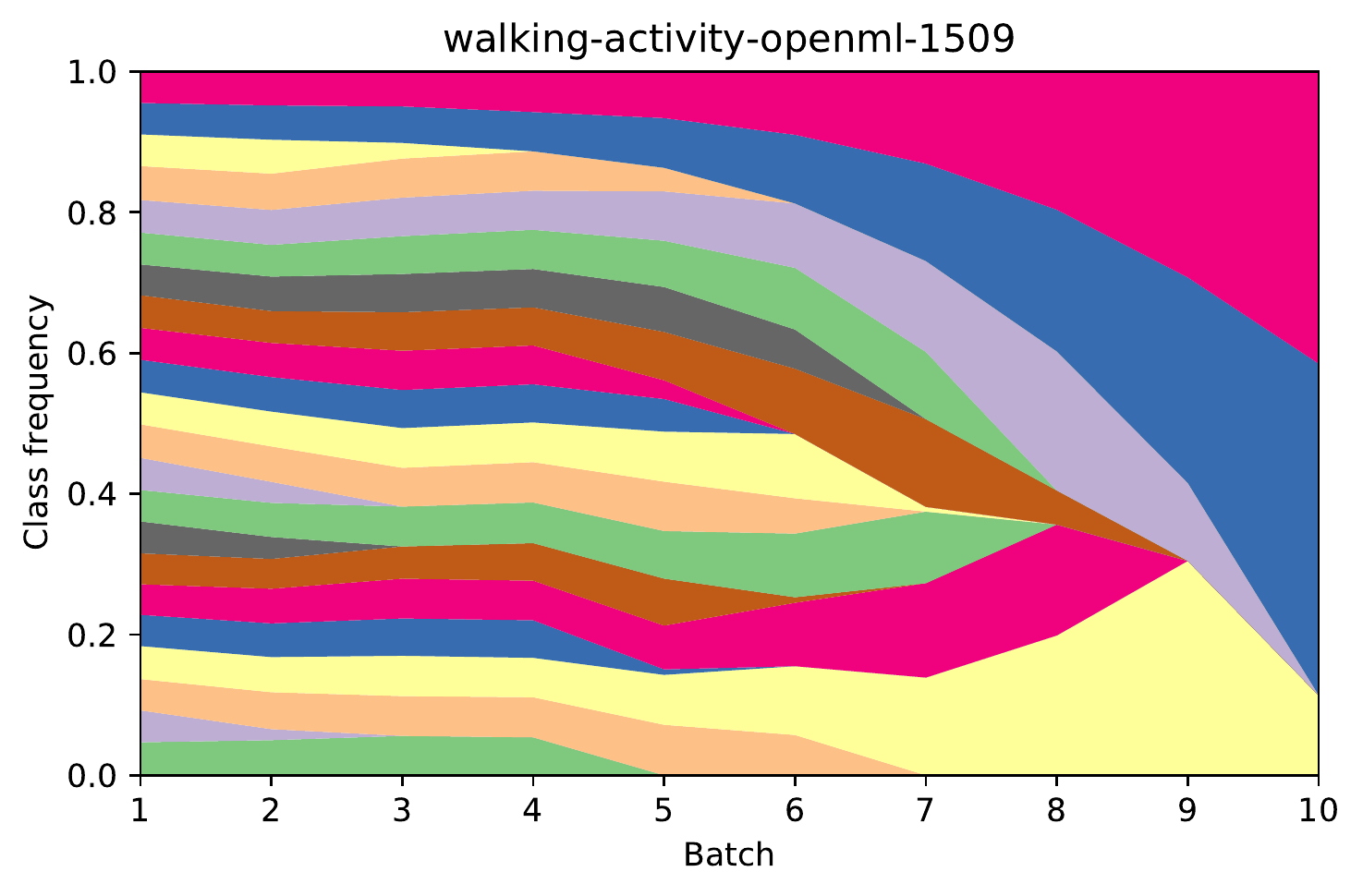}
  \includegraphics[width=.49\textwidth]{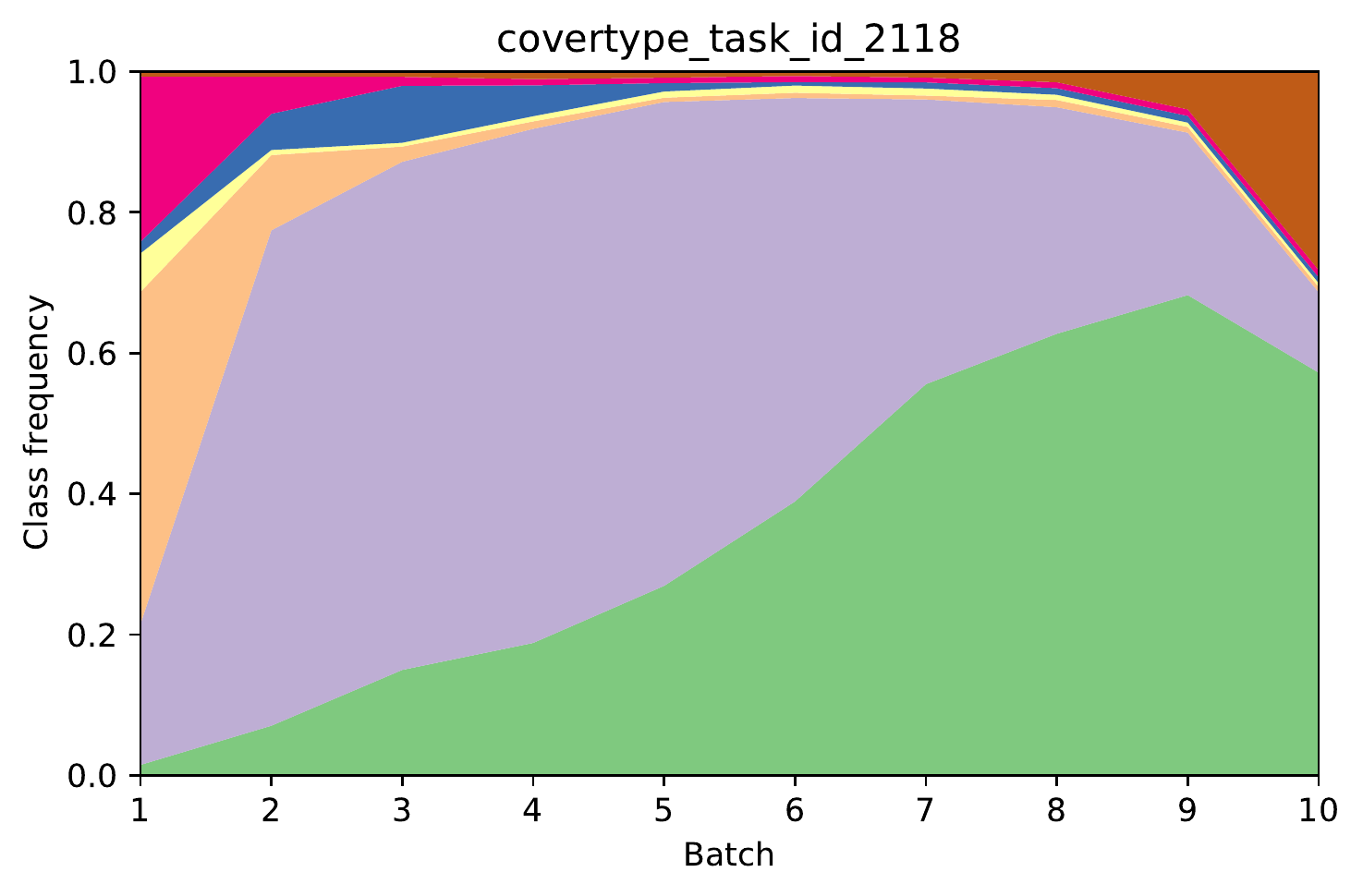}
  \caption{Class frequencies in the different batches of the task-free ``sorted''
  scenario.}
  \label{fig:class-frequencies}
\end{figure}

The task-free scenario for the tabular datasets is generated by sorting the
data points according to the value of a single feature.
We arbitrarily chose the first feature.
The resulting sequence is split into $10$ batches of approximately equal size.
Since the sequence changes smoothly, there is no notion of distinct tasks.
Nevertheless, the sorting generates a non-trivial pattern in the relative frequencies
of the classes across the $10$ batches, see Fig.~\ref{fig:class-frequencies}.

\subsection{Model architectures}

All experiments have been implemented using PyTorch \citep{paszke2019pytorch}.

\paragraph{MLP}
The MLP architecture consists of two fully-connected hidden layers with 128
units each and ReLU activation, followed by a fully-connected output layer.

\paragraph{CNN}
The CNN architecture consists of four convolution blocks, each comprising of a
convolutional layer with a $64$ filters with a receptive field of $3\times 3$ pixels
and padding of $1$ pixel, followed by batch normalization, ReLU activation, and
max pooling over $2\times 2$ windows.
The resulting feature map is average-pooled spatially, leading to a $64$-dimensional
representation.
This is followed by a fully-connected output layer.

\paragraph{ResNet-18} as described by \citet{he2016deep} and implemented in the
\texttt{torchvision} Python package.

\subsection{Optimization hyperparameters}

All models are trained using the Adam optimizer with a step size of $10^{-3}$
and default choices for the other hyperparameters.
We use a minibatch size of $100$.
No weight decay is applied.
We train each model for a fixed number of $200$ epochs.

\subsection{Parameters of GMC}

We use $s=4$ draws from the model's initialization distribution as implemented
by the standard initialization scheme in PyTorch.
For the random projections, we use $d=2000$.
This results in an embedding dimension of $D=sd = 8000$.
For the last-layer variant, the dimensions varies with the chosen model architecture
and dataset.

\section{Computational complexity of GMC}\label{a:compcomplex}

Orthogonal matching pursuit inverts a quadratic matrix of size $\vert I\vert$ in
each iteration.
This can be done efficiently by maintaining the Cholesky factorization of the matrix
$G_I^T G_I$ and updating it when a new element is added, see
\citet{rubinstein2008efficient}.
The computational complexity of OMP is
$O(DNn + Nn^2 + n^3)$, where $D$ is the dimension of the
gradient embedding, $N$ is the size of the original dataset and $n$ is the
desired coreset size (see, e.g., \citet{rubinstein2008efficient}).
In addition, we need $O(ND)$ memory to store the gradient embedding matrix.
While we can ``choose'' the gradient embedding dimension $D$, it needs to be
large enough to support the construction of a coreset of size $n$.
This means at least $D\geq n$; otherwise the matrix $G_I^T G_I$ will become
singular as soon as $\vert I \vert > D$.
Therefore, the algorithm also needs $O(Nn)$ memory.
Both the computational complexity and the memory requirements restrict the
applicability of the method to moderate coreset sizes.
It is quite common in the literature
to find experiments with corsets between 100 and 500 elements, but we show
that our method can scale to higher values and experiment with coresets sizes
up to 5000.

The cost of obtaining the gradient embeddings corresponds to $s$ epochs of
training on the full dataset.
In our experiments, we achieved good results with values as small as $s=4$.

\end{document}